\documentclass[journal]{IEEEtran}
\usepackage{latexsym}
\usepackage{amsmath,epsfig}
\usepackage{enumitem}
\usepackage{multirow}
\usepackage{subfigure}
\usepackage{multicol}
\usepackage{booktabs}
\usepackage{bigstrut}
\usepackage{array}
\usepackage{amssymb}
\usepackage{url}

\usepackage{float}
\usepackage{indentfirst}
\usepackage{times}
\usepackage{psfrag}
\usepackage{cite, xcolor}
\usepackage{textcomp}
\usepackage{multirow}
\usepackage{multicol}
\usepackage{stfloats}
\usepackage[breaklinks=true,bookmarks=false]{hyperref}

\ifCLASSINFOpdf

\else

\fi

\newcommand{\tabincell}[2]{\begin{tabular}{@{}#1@{}}#2\end{tabular}}
\begin{document}

\title{Understanding and Predicting the Memorability of Outdoor Natural Scenes}

\author{Jiaxin~Lu,
        Mai~Xu,~\IEEEmembership{Senior Member,~IEEE},
        Ren~Yang,
        and~Zulin~Wang,~\IEEEmembership{Member,~IEEE} 
\thanks{Jiaxin Lu, Mai Xu, Ren Yang and Zulin Wang are with Beihang University, 100191 Beijing, China. This work was supported by the NSFC projects 61876013, 61922009 and 61573037. Mai Xu is the corresponding author of this paper (e-mail: Maixu@buaa.edu.cn).}
}

%



\maketitle
\begin{abstract}
Memorability measures how easily an image is to be memorized after glancing, which may contribute to designing magazine covers, tourism publicity materials, and so forth. Recent works have shed light on the visual features that make generic images, object images or face photographs memorable. However, these methods are not able to effectively predict the memorability of outdoor natural scene images. To overcome this shortcoming of previous works, in this paper, we provide an attempt to answer: ``what exactly makes outdoor natural scenes memorable''. To this end, we first establish a large-scale outdoor natural scene image memorability (LNSIM) database, containing 2,632 outdoor natural scene images with their ground truth memorability scores and the multi-label scene category annotations. Then, similar to previous works, we mine our database to investigate how low-, middle- and high-level handcrafted features affect the memorability of outdoor natural scenes. In particular, we find that the high-level feature of scene category is rather correlated with outdoor natural scene memorability, and the deep features learnt by deep neural network (DNN) are also effective in predicting the memorability scores. Moreover, combining the deep features with the category feature can further boost the performance of memorability prediction. Therefore, we propose an end-to-end DNN based outdoor natural scene memorability (DeepNSM) predictor, which takes advantage of the learned category-related features.
Then, the experimental results validate the effectiveness of our DeepNSM model, exceeding the state-of-the-art methods. Finally, we try to understand the reason of the good performance for our DeepNSM model, and also study the cases that our DeepNSM model succeeds or fails to accurately predict the memorability of outdoor natural scenes. 

Our LNSIM dataset is available at \url{https://github.com/JiaxinLu-home/Natural-Scene-Memorability-Dataset}. The test code of the proposed DeepNSM method is publicly released at \url{https://github.com/RenYang-home/Natural-Scene-Memorability}.
\end{abstract}


%
\IEEEpeerreviewmaketitle

\section{Introduction}

One hallmark of human cognition is the splendid capacity of recalling thousands of different images, some in details, after only a single view \cite{standing1973learning, vogt2007long}. In fact, not all images are remembered equally in human brain. Some images stick in our minds, while others fade away in a short time \cite{standing1973learning, vogt2007long}. Memorability measures how easily an image is to be memorized after glancing. This kind of capacity is likely to be influenced by individual experiences, and is also subject to some degree of inter-subject variability, similar to some subjective image properties. This paper focuses on studying the image memorability of outdoor natural scenes, which frequently appear in human life. Understanding and predicting the memorability of outdoor natural scenes may have some potential applications. For example, we can use more memorable cover images for geography magazines or tourism advertisements to leave a deeper impression on readers, and we may chose a more impressive background or cover photo in social media profile to facilitate social communication, etc.


\begin{figure}[!t]
\begin{center}
\includegraphics[width=.9\linewidth]{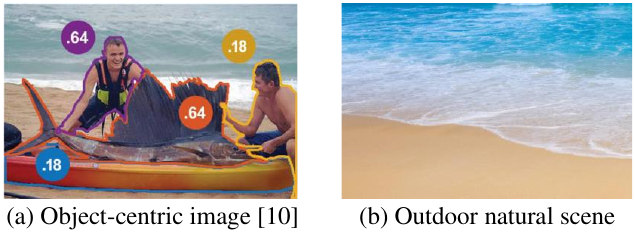}
\end{center}
\caption{An example of (a) object-centric image and (b) outdoor natural scene image. Note that the numbers in (a) indicate the memorability score of each object.}\label{fig:new1}
\end{figure}

Recent works \cite{isola2011makes,isola2014makes,khosla2012image,khosla2015understanding,isola2011understanding,mancas2013memorability,bylinskii2015intrinsic,dubey2015makes,baveye2016deep} analyze the fact that images are not equally remembered by humans, and provide reliable solutions for ranking images by memorability scores. These works are proposed for generic images. For example, as shown in Fig. \ref{fig:new1}-(a), Dubey \textit{et al.} \cite{dubey2015makes} explored the memorability of different objects in an image, and studied the contribution of each object to the memorability of the whole image.  In other words, according to \cite{dubey2015makes}, the features of the object compositions in a memorable image are different from those in an image that is hard to be memorized. However, understanding and predicting the memorability of the image background (beach and sea) in Fig. \ref{fig:new1}-(a) are neglected in \cite{dubey2015makes}. That is, the method of \cite{dubey2015makes} is not effective for the outdoor natural scene image\footnote{In this paper, we refer outdoor natural images as the natural images, which are without any salient object such as people, animals, and man-made objects.} shown in Fig. \ref{fig:new1}-(b), which is not an object-centric image, and has no obvious cue to clarify which part sticks in mind or fades away. Therefore, in this paper, we exclude the object-related features and explore the effective features, especially the macroscopic features describing the whole image, for predicting the memorability of outdoor natural scenes.

\begin{figure}[!t]
\vspace{-1em}
\begin{center}
\includegraphics[width=.99\linewidth]{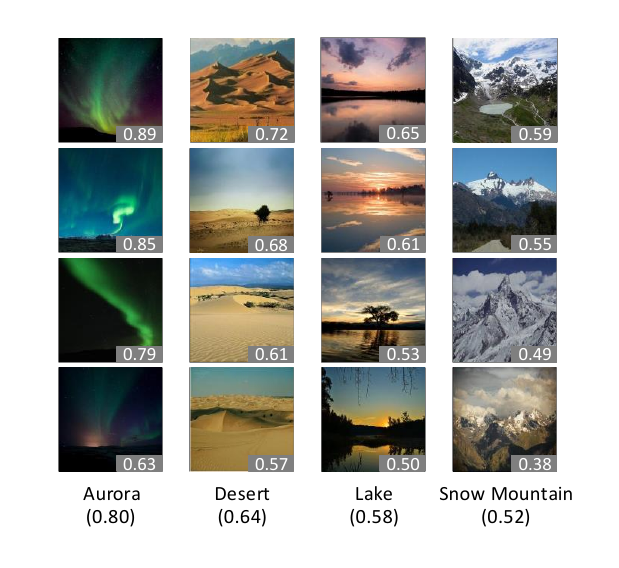}
\end{center}
\vspace{-2em}
\caption{Image samples from different outdoor natural scene categories. The ground truth memorability score is annotated in each image. The number in the bracket represents average score of each category.}\label{fig:Fig. first}
\vspace{-1em}
\end{figure}

Fortunately, Fig. \ref{fig:Fig. first} shows that scene category, as a high-level feature describing the content of the whole image, has correlation with the outdoor natural scene memorability. Moreover, deep neural network (DNN) is widely used in computer vision tasks, including predicting generic image memorability \cite{khosla2015understanding, baveye2016deep}. Inspired by this, we further investigate the influence of the scene category feature when combining with the baseline deep network, and further propose an end-to-end DNN method utilizing the category-related feature to predict the memorability of outdoor natural scenes. More importantly, thanks to the attempt for studying the memorability of images without salient objects, this paper advances the understanding of memorability in previous works. For example, we find that the word frequency might be the intrinsic factor making scene category an effective feature to predict memorability, since the most memorable categories have the least word frequencies. That is, the outdoor natural scenes, which people have rarely seen, e.g., aurora in Fig. \ref{fig:Fig. first}, tend to leave deep impression in mind.

Besides, the existing methods for generic images fail to accurately predict the memorability of outdoor natural scenes. For example, the method of \cite{isola2014makes} achieves $\rho=0.54$ on generic images, but only has $\rho=0.32$ on the outdoor natural subset of \cite{isola2014makes} and $\rho=0.33$ on the outdoor natural scene dataset LNSIM, established in this paper. Similarly, MemNet \cite{khosla2015understanding} and MemoNet \cite{baveye2016deep} reach the performance of $\rho=0.64$ and $\rho=0.636$ for generic images, but only have $\rho=0.43$ and $\rho=0.39$ on our LNSIM database for outdoor natural scenes. These also motivate our work.

This paper significantly extends our conference paper \cite{lu2018what}. Specifically, the extensions mainly include: (1) the discussion for our motivation, (2) the analysis on the influence of combining different non-deep features with deep features for predicting memorability (the motivation for designing our DeepNSM model), (3) understanding the effectiveness of scene category from the aspect of word frequency, (4) understanding how deep features work by investigating internal representation learnt by our DNN model, (5) the case study demonstrating the successful cases of the models utilizing non-deep features, deep features and our final model DeepNSM, and analyzing the failure cases of these three models. We also utilize the gray-level co-occurrence matrix (GLCM) to analyze the cases that our DeepNSM model is successful or unsuccessful for accurate memorability prediction, and (6) we add more experiments including utilizing more metrics to evaluate the prediction performance and the performance of the scene classifier. The main contributions of this paper are as follows.



(1) For understanding and predicting the memorability of outdoor natural scenes, we establish a large-scale outdoor natural scene database (LNSIM) with 2,632 outdoor natural images, which has 10 times number of images than the NSIM database \cite{Lu2016Predicting}. Also, the number of images in our LNSIM database is much more than the number of outdoor natural scene images in generic datasets \cite{isola2011makes,isola2014makes, khosla2015understanding, baveye2016deep}. In our LNSIM database, all images are with the memorability scores, obtained from the memory game with more than 100 volunteers. Differing from other previous works, all images in our database have multi-label scene category annotations, for investigating the relationship between scene category and the memorability of outdoor natural scenes.

(2) Based on the previous works \cite{isola2011makes,isola2014makes, khosla2015understanding, isola2011understanding,bylinskii2015intrinsic} for memorability of generic images, our work further
thoroughly study the relationship between outdoor scene memorability and various kinds of features, including the deep features and the handcrafted features of low-, middle- and high-levels. Most importantly, we investigate the influence of all handcrafted features on combining with DNN to predict outdoor natural scene memorability. We not only observe that the scene category has the highest correlation with outdoor natural scene memorability among different levels of handcrafted features, and also find that combining the scene category feature with our baseline deep model boosts the performance of DNN.

(3) In accordance with the above findings, we propose a DNN based outdoor natural scene memorability (DeepNSM) predictor, which integrates category-related features with a conventional DNN for memorability prediction on outdoor natural scene images. It differs from the previous works \cite{isola2011makes,isola2014makes, khosla2015understanding, isola2011understanding,bylinskii2015intrinsic}, which either solely apply handcrafted visual features or utilize a simple DNN (e.g., applying GoogleNet in \cite{baveye2016deep}). It is worth pointing out that our DeepNSM approach automatically extracts category-related features by DNN and does not need any manual category annotation when predicting memorability.

(4) Moreover, we investigate the word frequency of each scene category, and find that the images which are easy to be memorized are with low frequency, i.e., they rarely appear in human daily life. This is probably because these images more easily leave deep impression in people's mind. Then, we provide an attempt to understand how our DeepNSM model works to predict memorability by visualizing the internal representation of the last convolutional layer, and observe that our DeepNSM model is able to locate the memory region more accurately than the compared methods, thus leading to better performance.

(5) Finally, we demonstrate the cases that the category feature, deep features and our DeepNSM model are effecitve for memorability prediction on outdoor natural scenes, and also study the cases that they fail to accurately predict memorability.
To understand the successful and failure cases, we further analyze the texture of images, and find that the memorability of the images with higher contrast, lower homogeneity and lower intra-picture correlation are more likely to be predicted accurately by our DeepNSM method.

\begin{figure*}[t]
\begin{center}
\includegraphics[width=1\linewidth]{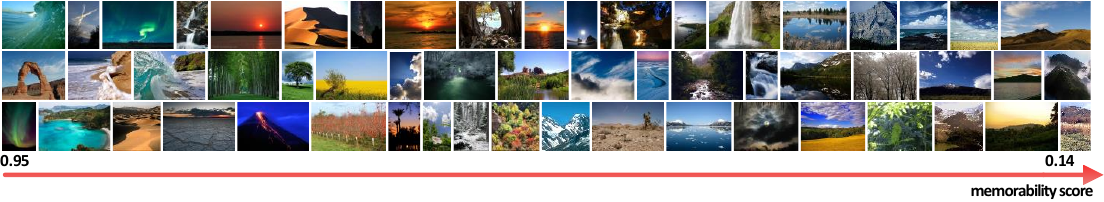}
\end{center}
\vspace{-1em}
\caption{Image samples from our LNSIM database. The images above are ranked by their memorability scores, which decrease from left to right.}\label{fig:Figure 1}
\end{figure*}

\section{Related work}\label{related}

\textbf{Memorability of generic images.} Isola \textit{et al.} \cite{isola2011makes} pioneered on the study of image memorability for generic images, and they have shown that memorability is an intrinsic property of an image. They further analyzed how various visual factors influence the memorability of generic images, and a greedy feature-selection scheme was employed to obtain the best feature combination for memorability prediction \cite{isola2011understanding}. The global features (i.e., pixels, GIST \cite{oliva2001modeling}, SIFT \cite{lazebnik2006beyond}, HOG2$\times$2 \cite{felzenszwalb2010object,dalal2005histograms}) were verified and then combined. As a result, the accurate prediction performance of memorability is achieved on generic images \cite{isola2011makes,isola2011understanding,isola2014makes}. Besides, Khosla \textit{et al.} \cite{khosla2012image} combined local features with global features to increase the prediction performance. Later, Bylinskii \textit{et al.} \cite{bylinskii2015intrinsic} investigated the interplay between intrinsic and extrinsic factors that affect image memorability, and then they developed a more complete and fine-grained model for image memorability.

In addition to the hand-crafted features, visual attention has  recently received significant research interests, such as visual attention prediction \cite{wang2019salient} and human visual behavior modeling \cite{wang2019revisiting, wang2019inferring}. Inspired by the studies on visual attention, Mancas \textit{et al.} \cite{mancas2013memorability} suggested that incorporating the attention-related feature in \cite{isola2011makes} further improves the prediction accuracy.
Meanwhile, a visual attention-driven approach was proposed in \cite{celikkale2013visual}.

Recently, DNN is utilized on image memorability. For example, Siarohin \textit{et al.} \cite{siarohin2019increasing} proposed applying a neural style transfer algorithm to increase image memorability. Moreover, for predicting image memorability, several DNN approaches were proposed to significantly improve the prediction accuracy. Specifically, Khosla \textit{et al.} \cite{khosla2015understanding} trained the MemNet on a large-scale database, achieving a splendid prediction performance close to the human consistency. In addition, Baveye \textit{et al.} \cite{baveye2016deep} fine-tuned the GoogleNet on the same database of \cite{isola2011makes}, exceeding the performance of handcrafted features mentioned above. They also cast light on the importance of balancing emotional bias, when establishing the memorability-related database. However, the works of \cite{khosla2015understanding} and \cite{baveye2016deep} only re-trained the single DNNs (e.g., AlexNet and GoogleNet), and did not take advantage of any other features for predicting image memorability.

\textbf{Memorability of faces, objects and outdoor natural scene.} To better understand and predict image memorability, the study of image memorability on certain targets, like faces, objects and outdoor natural scenes, has recently attracted the interests of computer vision researchers \cite{khosla2013modifying,bainbridge2012establishing,bainbridge2013intrinsic,dubey2015makes,Lu2016Predicting}. Bainbridge \textit{et al.} \cite{bainbridge2012establishing} firstly established a database for studying the memorability of human face photographs. They further explored the contribution of certain traits (e.g., kindness, trustworthiness, etc.) to face memorability, but such traits only partly explain facial memorability.
Furthermore, \cite{khosla2013modifying} proposed a method to modify the memorability of individual face photographs.

Dubey \textit{et al.} \cite{dubey2015makes} were the first to study the problem of object memorability. They assumed that object categories play an important role in determining object memorability; therefore, they obtained the memorability scores of all constituent objects possibly appearing in object images by subjective experiment. Since the splendid performance of DNN is achieved in various recognition tasks, Dubey \textit{et al.} \cite{dubey2015makes} utilized the deep features extracted by conv-net \cite{krizhevsky2012imagenet,jia2014caffe} and ground truth scores of objects to predict object memorability better.
It is worth pointing out that although object category and DNNs were used in \cite{dubey2015makes}, they did not design an end-to-end deep network for memorability prediction. In the DL-MCG method of \cite{dubey2015makes}, object segments are generated by using a DNN, and then they trained a support vector regressor (SVR) to map deep features to memorability scores. Thus, it is not able to be optimized in an end-to-end manner. Besides, Dubey \textit{et al.} \cite{dubey2015makes} also provided an upper bound of their DL-MCG method, where the ground-truth object segmentation is used to replace the DNN before SVR. Such upper bound has to be achieved with manual annotation of segmentation and without using DNNs. In this paper, our DeepNSM approach, as an end-to-end DNN, is able to automatically extract category-related feature, and to predict the memorability of outdoor natural scenes without using any manual annotation. Separately, \cite{han2015learning} learned video memorability from brain functional magnetic resonance imaging (fMRI).

More recently, Lu \textit{et al.} \cite{Lu2016Predicting} studied the memorability of outdoor natural scene on the subset of database in \cite{isola2011makes}. They indicated that the HSV color features perform well on the outdoor natural scene memorability, and then they combined the HSV-based feature and other traditional low-level features to predict memorability scores. Nonetheless, only handcrafted low-level features are considered in memorability prediction, which are limited in the prediction accuracy.

\begin{figure*}[t]
\begin{center}
\includegraphics[width=\linewidth]{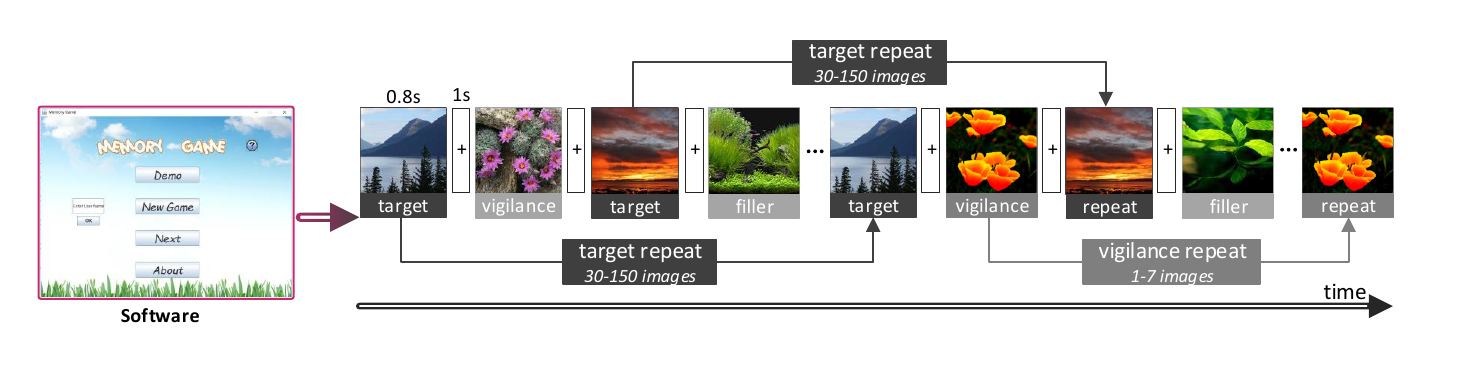}
\end{center}
\caption{The experimental procedure of our memory game. Each level lasts about 5.5 minutes with a total of 186 images. Those 186 images are composed of 66 targets, 30 fillers and 12 vigilance images. The specific time durations for experiment setting are labeled above.}\label{fig:game}
\end{figure*}

\section{Outdoor natural scene memorability database}\label{data}

As a first step towards understanding and predicting the memorability of outdoor nature scenes, we build the LNSIM database. Our LNSIM database is specifically established for the memorability of outdoor natural scenes, which is more than 10 times larger than the previous NSIM database \cite{Lu2016Predicting}.
In our LNSIM database, we follow \cite{isola2011makes} and \cite{isola2014makes} to utilize the memory game with 104 volunteers for obtaining the ground-truth memorability score of each image.
Besides, scene category is an essential feature of outdoor natural scenes, and may contribute to predicting the memorability. Therefore, we also annotate each image in our LNSIM database with scene category label.


\textbf{Collecting images.} In our LNSIM database, there are in total 2,632 outdoor natural scene images. For obtaining these images, we first selected 6,886 images, which contain outdoor natural scenes from the existing databases, including MIR Flickr \cite{Huiskes2008The}, MIT1003 \cite{Judd2009Learning}, NUSEF \cite{NUSEF2010}, SUN \cite{Xiao2010}, affective image database \cite{Machajdik2010Affective}, and AVA database \cite{Murray2012AVA}. Since the outdoor natural scene images are hard to be distinguished, 5 volunteers were asked to select the outdoor natural scene images from 6,886 images with the following two criteria \cite{Lu2016Predicting}:

(1) Each image is with outdoor natural scenes.

(2) Each image is only composed of outdoor natural scenes, not having any human, animal and man-made object.

\begin{figure}[!t]
\begin{center}
\includegraphics[width=\linewidth]{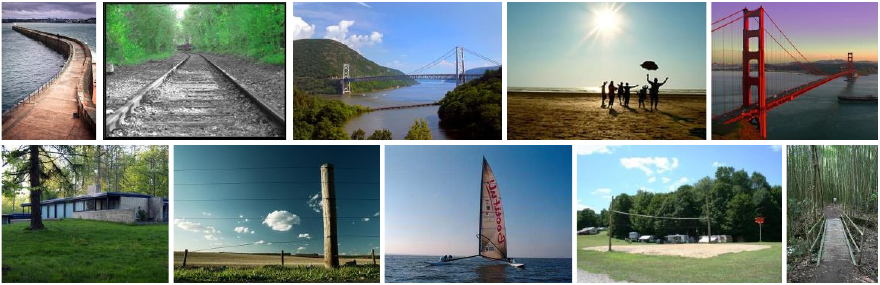}
\end{center}
\vspace{-1em}
\caption{Examples of candidate images which are not selected to our LNSIM database. It can seen that these images contain humans or man-made objects, thus not matching our definition of outdoor natural scene images.}\label{fig:eg}
\end{figure}

Afterwards, the images, chosen by at least four volunteers, were included in our LNSIM database. As a result, 2,632 outdoor natural scene images were obtained for the LNSIM database, to be scored with memorability. Note that the resolution of these images ranges from 238$\times$168 to 3776$\times$2517. Fig. \ref{fig:Figure 1} shows some example images from our LNSIM database, and Fig. \ref{fig:eg} illustrates the images which are not selected from the 6,880 candidates.

\textbf{Memorability scores.} In our experiment, we set up a memory game, which was used to quantify the memorability of each image in our LNSIM database. The memory game is crowdsourced with totally 104 volunteers (47 females and 57 males). They do not overlap with the volunteers who participated in the image selection. As shown in Fig. \ref{fig:game}, the procedure of our memory game is similar with that in \cite{khosla2015understanding}. Note that compared to \cite{khosla2015understanding}, more time is allowed for a subject to decide whether the image has been seen before. The reason is that it normally takes more time for human to memorize outdoor natural scenes for lack of salient objects \cite{isola2014makes}.

%
%

In our experiment, there were 2,632 target images, 488 vigilance images and 1,200 filler images, which were unknown to all subjects. Vigilance and filler images were randomly sampled from the rest of 6,886 images. Target images, as stimuli for our experiment, were randomly repeated with a spacing of 35-150 images. Vigilance images were repeated within 7 images, in attempt to ensure that the subjects were paying attention to the game. Filler images were presented once, such that spacing between the same target or vigilance images can be inserted.
%

On average, we obtained over 80 valid memory results per target image. The average hit rate on target images was 73.7\% with standard deviation (SD) of 14.2\%, running on the experimental results of 104 subjects. Compared with the database of generic images (average score: 67.5\%, SD: 13.6\%), this implies that the subjects indeed concentrated on the game. The average false alarm rate was 8.14\% (SD of 0.81\%). As the false alarm rate was low in comparison with the hit rate, it eliminates the possibility of hitting correct images only by chance. Thus, our data can reliably reflect the memorability of outdoor natural scene images.
After collecting the data, we assigned a memorability score to quantify how memorable each image is, following the way of \cite{khosla2015understanding}. Since the time intervals of repeat on target images were various in our experiment, we follow the method of [6] to regularize the various time intervals to a certain time $T$. In this paper, we set $T$ to be the time duration of displaying 100 images, as the repeat spacing of targets ranges from 35 to 150.

\begin{figure}[t]
\begin{center}
\includegraphics[width=0.8\linewidth]{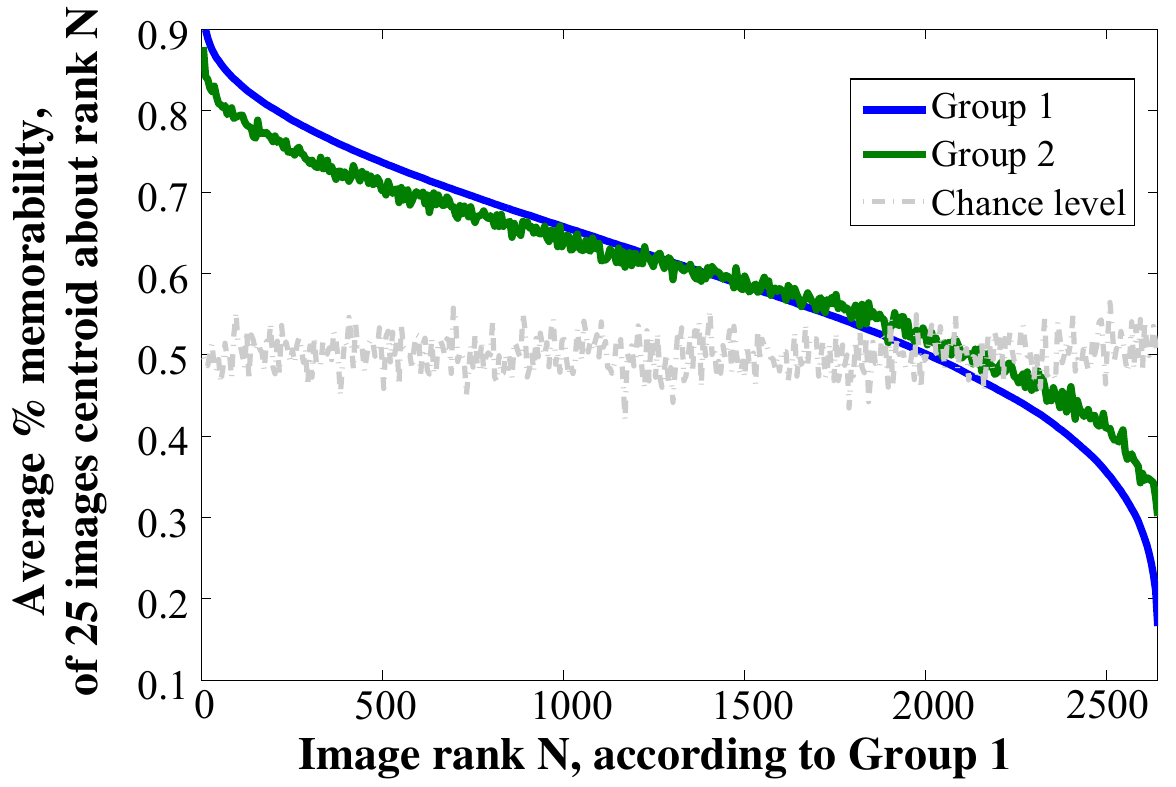}
\end{center}
\caption{Measure of human consistency in outdoor natural scene memorability. The memorability scores are derived from two groups of subjects. Images are ranked by memorability scores of subjects in Group 1, and then the curves plot the average memorability scores of Group 1 vs. Group 2. For clarity, we convolve the resulting plots with a length-6 box filter along with the horizontal axis. The chance line is provided by allocating random prediction scores as a reference.}\label{fig:img3}
\end{figure}

\begin{table*}[!t]
\small
  \centering
  \caption{The numers of images in each scene category.}
    \begin{tabular}{|l|c||l|c||l|c||r|r|}
    \hline
    \multicolumn{1}{|c|}{\textbf{Category}} & \textbf{Image no.} & \multicolumn{1}{c|}{\textbf{Category}} & \textbf{Image no.} & \multicolumn{1}{c|}{\textbf{Category}} & \textbf{Image no.} & \multicolumn{1}{c|}{\textbf{Category}} & \multicolumn{1}{l|}{\textbf{Image no.}} \\
    \hline
    arch  & 33    & desert\_road & 54    & ice\_shelf & 11    & \multicolumn{1}{l|}{river} & \multicolumn{1}{c|}{261} \\
    \hline
    aurora & 4     & dew   & 3     & islet & 38    & \multicolumn{1}{l|}{rock\_arch} & \multicolumn{1}{c|}{20} \\
    \hline
    badlands & 121   & farm  & 58    & lagoon & 80    & \multicolumn{1}{l|}{ruin} & \multicolumn{1}{c|}{10} \\
    \hline
    bamboo\_forest & 12    & field-cultivated & 131   & lake-natural & 271   & \multicolumn{1}{l|}{sea} & \multicolumn{1}{c|}{353} \\
    \hline
    beach & 123   & field-wild & 275   & lawn  & 144   & \multicolumn{1}{l|}{sky} & \multicolumn{1}{c|}{1401} \\
    \hline
    beach\_house & 6     & field\_road & 61    & lighthouse & 5     & \multicolumn{1}{l|}{snowfield} & \multicolumn{1}{c|}{113} \\
    \hline
    boat\_desk & 42    & fishpond & 1    & lightning & 6     & \multicolumn{1}{l|}{star} & \multicolumn{1}{c|}{47} \\
    \hline
    botanical\_garden & 24    & flower & 127   & marsh & 109   & \multicolumn{1}{l|}{sun} & \multicolumn{1}{c|}{216} \\
    \hline
    butte & 131   & forest & 669   & moon  & 46    & \multicolumn{1}{l|}{swamp} & \multicolumn{1}{c|}{44} \\
    \hline
    canal-natural & 24    & forest\_path & 95    & mountain & 656   & \multicolumn{1}{l|}{topiary\_garden} & \multicolumn{1}{c|}{4} \\
    \hline
    canyon & 52    & forest\_road & 19    & mountain\_cloudy & 81    & \multicolumn{1}{l|}{tree} & \multicolumn{1}{c|}{378} \\
    \hline
    cliff & 115   & glacier & 13    & mountain\_path & 65    & \multicolumn{1}{l|}{tree\_farm} & \multicolumn{1}{c|}{32} \\
    \hline
    coast & 287   & golf\_course & 9     & mountain\_snowy & 198   & \multicolumn{1}{l|}{tree\_hole} & \multicolumn{1}{c|}{4} \\
    \hline
    creek & 128   & grotto & 27    & pasture & 20    & \multicolumn{1}{l|}{valley} & \multicolumn{1}{c|}{70} \\
    \hline
    crevasse & 4     & hayfield & 7     & pier  & 14    & \multicolumn{1}{l|}{volcano} & \multicolumn{1}{c|}{10} \\
    \hline
    dam   & 2     & highway & 24    & pond  & 94    & \multicolumn{1}{l|}{waterfall} & \multicolumn{1}{c|}{54} \\
    \hline
    desert-sand & 121   & iceberg & 4     & rainbow & 25    & \multicolumn{1}{l|}{wave} & \multicolumn{1}{c|}{138} \\
    \hline
    desert-vegetation & 75    & ice\_floe & 19    & rainforest & 121   &       &  \\
    \hline
    \end{tabular}%
  \label{tab:no}%
\end{table*}%

\textbf{Human consistency.} Next, we follow \cite{isola2011makes} and \cite{isola2014makes} to quantify the human-to-human consistency of memorability over our LSNIM database. To evaluate human consistency, we randomly split subjects into two independent halves, and then measured the correlation between memorability scores of these two halves. We examined consistency with a variant of correlation measurement: We sorted all the 2,632 images by their scores of the first half of subjects, and calculated the corresponding cumulative average memorability scores, according to the second half of subjects. In this fashion, Fig. \ref{fig:img3} plots the memorability scores measured by independent two sets of subjects and averaged over 25 random splits, in which the scores of Group 1 are set as benchmark. Note that the horizontal axis ranks the images with the memorability scores of the first half (denoted as Group 1) in the decreasing order. As shown in Fig. \ref{fig:img3}, there exists high consistency between two groups of subjects, especially compared to that of the random prediction.

We further quantified the human-to-human consistency by measuring the Spearman rank correlation coefficient (SRCC, denoted by $\rho$). The SRCC on the LSNIM database is 0.78 between two sets of scores measured over 25 random splits. Compared with \cite{isola2014makes}, SRCC measured on outdoor natural scene image set is a little higher than that calculated in generic image set ($\rho=0.75$). Furthermore, we selected the top 100 most memorable images with average score of 83.5\% marked by Group 1, and then we obtained an average score of 79.4\% from second half of subjects (denoted as Group 2). The above results indicate that the individual differences add noise to estimation; nonetheless, different subjects tend to remember or forget the same images.

To conclude, humans are highly consistent in remembering outdoor natural scene images. This replicates prior findings in \cite{isola2011makes, isola2014makes} that the memory game generates effective ground truth memorability scores.
This also indicates that the memorability of outdoor natural scenes can potentially be predicted with high accuracy. 
Note that, as aforementioned, the memorability scores of our LNSIM database are annotated by 104 volunteers, and we do not use the Amazon Mechanical Turk (MTurk) workers as in \cite{isola2011makes} and \cite{isola2014makes}. However, the human consistency of our LNSIM database ($\rho=0.78$) is comparable to that of \cite{isola2011makes} and \cite{isola2014makes} ($\rho=0.75$). Besides, it is worth pointing out that the slight difference between $\rho=0.75$ and $\rho=0.78$ may be due to the difference of image sets and participants of the memory games.

\textbf{Scene category labels.} According to WordNet taxonomy \cite{miller1995wordnet}, our LNSIM database includes 71 scene categories (badlands, coast, desert, etc.), which are non-overlapped with each other. The names of the 71 scene categories are shown in Fig. \ref{fig:scenecat} in Section IV. Note that, in our LNSIM database, each image may belong to multiple categories, i.e., it is a multi-label scene category database. To obtain the ground truth of scene category, we follow \cite{zhou2017places} to conduct two experiments to annotate the 2,632 scene category images in our database.

\begin{itemize}
\item Task 1 (Classification Judgment): We asked 5 participants to indicate which scene categories an image has. A random image query was generated for each participant. We showed an image and all scene categories at a time. Participants had to choose proper scene category labels to interpret scene stuff for each image.
\item Task 2 (Verification Judgment): We further ran a separate task on the same set of images by recruiting another 5 participants after Task 1. For a given category name, a single image was shown centered in the screen, with a question like ``\textit{is this a coast scene?}'' The participants were asked to provide a binary answer to the question for each image. The default answer was set to ``No'', and the participants can check the box of image index to set ``No'' to ``Yes''.
\end{itemize}

We annotated all images with categories through the majority voting over Task 1 and Task 2. Specifically, Task 1 completed the outdoor natural scene category annotation initially, while Task 2 amended the results of Task 1. For each image of our database, we determined its scene categories according to the results of Task 2. In this way, the scene categories of all 2,632 images in the LNSIM database were obtained, taking account of 10 participants' selection. Note that all 10 participants did not attend the memory game, and one image may have more than one category in our database. Additionally, the rate of choosing ``Yes'' in Task 2 is $81\%$ among the 5 participants. This indicates that different annotators are consistent in classifying scene category. Table \ref{tab:no} shows the names of the 71 categories and the image numbers of each category.

\textbf{Training and test sets.} In this paper, we refer to the scores collected by the aforementioned memory game as the ``ground truth'' memorability for each image, and refer to the multi-label annotation as the ``ground truth'' scene category of each outdoor natural scene. The 2,632 outdoor natural scene images with their ground truth memorability scores and category labels are randomly divided into the non-overlapping training and test sets. The training set contains 2,200 images, and the remaining 432 images are used for test.



\section{Analysis on outdoor natural scene memorability}\label{analysis}


In this section, we mine our LNSIM database to better understand how outdoor natural scene memorability is influenced by the low-, middle- and high-level handcrafted features.

\subsection{Low-level feature vs. memorability}

On the basis of predecessors \cite{isola2011makes,isola2014makes,khosla2015understanding}, it has been verified that low-level features, like pixels, SIFT \cite{lazebnik2006beyond} and HOG2$\times$2 \cite{felzenszwalb2010object,dalal2005histograms}, have impact on memorability of generic images. Here, we investigate whether these low-level features still work on purely outdoor natural scene image set as well. To this end, we train an SVR for each low-level feature on our training set for memorability prediction, and then evaluate the SRCC of these low-level features with memorability on the test set. The histogram intersection kernels\footnote{Note that we traverse all possible kernels for each feature, and select the one with the best performance.} are utilized for these features.
Note that, these low-level features are extracted in the same manner as \cite{khosla2012memorability}.

\begin{table}[t]
\scriptsize
  \centering
    \caption{The correlation $\rho$ between low-level features and outdoor natural scene memorability.}\label{tab:globalfeats}
    \begin{tabular}{cccccc}
    \toprule
  Database       &pixels&SIFT \cite{lazebnik2006beyond}&HOG\cite{felzenszwalb2010object}&Combination&Human\\
    \midrule
    Our LNSIM    & 0.08    & 0.28 & 0.29 & 0.33 & 0.78 \\
    \midrule
    Generic images \cite{isola2014makes}     & 0.22    & 0.41 & 0.43 & 0.45 & 0.75 \\
    \bottomrule
    \end{tabular}
\end{table}

Table \ref{tab:globalfeats} reports the results of SRCC on outdoor natural scenes, with SRCC on generic images \cite{isola2014makes} as the baseline. It is evident that pixels ($\rho=0.08$), SIFT ($\rho=0.28$) and HOG2$\times$2 ($\rho=0.29$) are not as effective as expected on outdoor natural scene images, especially compared with generic images. For example, the feature of SIFT has capacity to reflect the memorability of generic images to a certain degree with $\rho=0.41$, but its SRCC decreases to 0.28 on outdoor natural scene images. This suggests that the low-level features have decent performance on predicting the memorability of generic images; however, they cannot effectively characterize the visual information for remembering outdoor natural scenes. Then, we additionally train an SVR on a kernel sum of these low-level features, achieving a rank correlation of $\rho=0.33$. This is a bit far from the SRCC result ($\rho=0.45$) of feature combination for generic images.

Moreover, color is another low-level feature, which is analyzed in \cite{isola2011makes,isola2014makes} and utilized in \cite{Lu2016Predicting} for outdoor natural scene memorability prediction. To address whether it still works, we calculate the mean and variance of each HSV color channel for all images in our database. We also measure the SRCC of each color channel with the corresponding memorability scores.
As reported in \cite{Lu2016Predicting}, the HSV-based feature reaches $\rho=0.27$ on their small scale (258 images) NSIM database \cite{Lu2016Predicting}. Nevertheless, on our LNSIM database, which is more than 10 times larger than the NSIM database, the HSV-based feature only has the SRCC of $\rho=0.10$. This indicates that the color features cannot well explain outdoor natural scene memorability.

Besides, we also evaluate the p-values \cite{westfall1993resampling} of the predicted memorability by these low-level features. The p-values for the features of pixels, SIFT and HOG are $0.1053$, $3.8493\times10^{-9}$ and $7.3470\times10^{-10}$, respectively. The combination of these features has the p-value of $4.3129\times10^{-12}$. For HSV features, the p-values for the mean of H, S, and V channels are 0.0494, 0.6999, 0.7814, respectively. The p-values for the variance of H, S, and V channels are 0.5609, 0.2830, $8.6898\times10^{-4}$, respectively. Note that only if the p-value is less than $1.67\times 10^{-3}$, the predicted values are statistically significant\footnote{In this paper, there are totally 30 statistical experiments conducted, including low-, middle-, high-level features, deep features, combination of deep features and other features, the DeepNSM model and the four compared methods. Therefore, according to the Bonferroni Correction, the threshold of p-value is set to $0.05/30 \approx 1.67\times 10^{-3}$.}. Therefore, these further verify that directly using pixel values/colors is not able to effectively predict the memorability.




\subsection{Middle-level feature vs. memorability}

The middle-level feature of GIST \cite{oliva2001modeling} describes the spatial structure of an image. Previous work \cite{isola2011makes,isola2014makes} mentioned that GIST is correlated with memorability on generic images ($\rho=0.38$, see Table \ref{tab:midfeats}). In view of this observation, we train an SVR predictor with a RBF kernel for quantifying the correlation between the GIST feature and memorability of outdoor natural scenes. Note that the training set is used to tune the hyper-parameters for the kernels. Table \ref{tab:midfeats} shows that the SRCC of GIST is 0.23, much less than $\rho=0.38$ of generic images. This illustrates that structural information provided by the GIST feature is less effective for predicting memorability scores on outdoor natural scenes.

\begin{table}[t]
\scriptsize
  \centering
    \caption{The correlation $\rho$ between middle-level features and outdoor natural scene memorability.}\label{tab:midfeats}
    \begin{tabular}{cccccc}
    \toprule
  Database       &GIST &PQFT &SalGAN &DVA&SALICON\\

  &\cite{oliva2001modeling}& \cite{guo2010novel}& \cite{pan2017salgan}& \cite{wang2018deep}&\cite{jiang2015salicon}\\

    \midrule
    Our LNSIM     & 0.23  & 0.25  & 0.20 & 0.20 & 0.14 \\
    \midrule
    Generic images \cite{isola2014makes}     & 0.38    & 0.15  & 0.27  &  0.30 & 0.16 \\
    \bottomrule
    \end{tabular}
\end{table}

\begin{figure}[t]
\begin{center}
\includegraphics[width=0.88\linewidth]{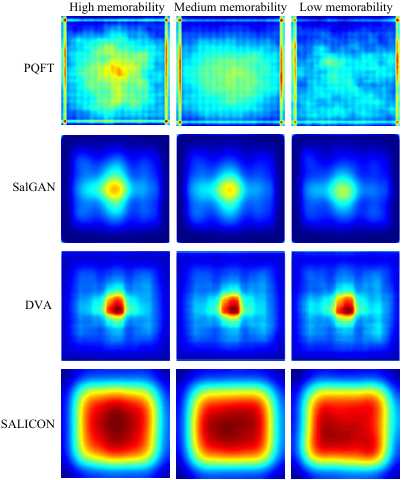}
\end{center}
\caption{Averaged saliency maps on images of high, medium and low memorability in our LNSIM database.}\label{fig:sa}
\end{figure}

\begin{figure*}[t]
\begin{center}
\subfigure{\includegraphics[width=0.80\linewidth]{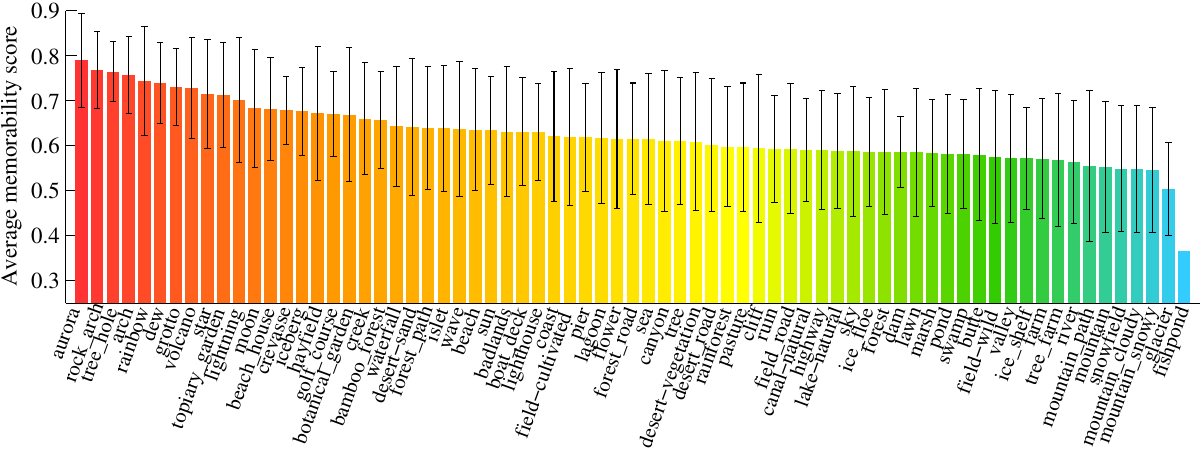}}
\end{center}
\caption{Comparison of average memorability score and standard deviation of each scene category.}\label{fig:scenecat}
\end{figure*}

\begin{table*}[!t]
  \centering
  \footnotesize
    \caption{The SRCC ($\rho$) of the combination of different hand-crafted features with the deep feature.}\label{tab:deep}
    \begin{tabular}{|c|c|c|c|c|c|c|c|c|c|c|}
    \hline
     \multirow{2}[2]*{} & & \multicolumn{4}{c|}{DF\ +\ low-level}&\multicolumn{4}{c|}{DF\ +\ middle-level} & DF+high-level \\
     \hline
     & DF & DF+pixels & DF+SIFT & DF+HOG & DF+HSV & DF+GIST & DF+PQFT & DF+SalGAN & DF+DVA & DF+category\\
    \hline
    $\rho$  &  0.4406 & 0.4431 & 0.4430 & 0.4420 & 0.4407 & 0.4413 & 0.4419 & 0.4413 & 0.4440 & \textbf{0.4610} \\
    \hline
    \end{tabular}
\end{table*}

Intuitively, the region that attracts visual attention \cite{guo2010novel,pan2017salgan,wang2018deep, wang2018deep2,wang2017deep} in an outdoor natural scene may affect image memorability. The work of \cite{mancas2013memorability,celikkale2013visual} attempted to explain memorability of generic images using visual attention-driven features. To quantify the correlation of visual attention with memorability on outdoor natural scenes, we apply four state-of-the-art models of visual attention (i.e., PQFT \cite{guo2010novel}, SalGAN \cite{pan2017salgan}, DVA \cite{wang2018deep} and SALICON \cite{jiang2015salicon}) to extract saliency maps.
Similar to other features, we utilize an SVR predictor to measure the SRCC of the saliency features. Note that the RBF kernel is chosen for the saliency features. We further split our LNSIM database into three classes: high memorability (score $\geq0.7$), medium memorability ($0.7>$ score $\geq0.4$) and low memorability (score $<0.4$). Fig. \ref{fig:sa} demonstrates the averaged saliency maps of each class.

Additionally, Table \ref{tab:midfeats} compares the correlation of saliency features with memorability of outdoor natural scene images and generic images. It can be seen from Table \ref{tab:midfeats} that SALICON \cite{jiang2015salicon}, which is the best saliency detection method among the four methods, has the least correlation with outdoor natural scene memorability. However, PQFT \cite{guo2010novel}\footnote{As Fig. \ref{fig:sa} shows, there are some boundary artifacts in the PQFT saliency maps. To reduce the boundary artifacts, we crop the boundary or weight the saliency map with a Gaussian filter to reduce the boundary artifacts. However, this leads to performance degradation of memorability prediction. Hence, we keep the boundary in the saliency maps generated by the PQFT method.}, which has the worst saliency detection performance, is the most effective one for predicting the memorability of outdoor natural scene images. Such conclusion can also be found from Fig. \ref{fig:sa}. These results indicate that the effectiveness of the saliency feature for predicting memorability is not correlated with the saliency detection accuracy.
Besides, this also suggests that when predicting outdoor natural scene memorability, frequency domain saliency model (PQFT) performs better than other pixel domain models.


\subsection{High-level feature vs. memorability}

There is no salient object, animal or person in outdoor natural scenes, such that scene category, as a high-level feature, may be effective to interpret outdoor natural scene images.
Similar to object detection, we use scene category attribute to characterize scene semantics of each outdoor natural scene image. For generic images, Bylinskii \textit{et al.}, \cite{bylinskii2015intrinsic} showed that the memorability has correlation with object cattery. In the following, we explore the relationship between the memorability of outdoor natural scenes and the scene category.

As mentioned in Section III, we annotated each image with scene category labels in our LSNIM database. Now, we test the memorability prediction performance of scene category on the LNSIM database. An SVR predictor with the histogram intersection kernel is trained for scene category. The scene category attribute achieves a good performance of SRCC ($\rho=0.38$, p-value $=2.4516\times10^{-14}$), outperforming the results of low- and middle-level features. This suggests that scene category, as a high-level feature, is an obvious cue of quantifying the outdoor natural scene memorability. We further analyze the connection between different scene categories and outdoor natural scene memorability. To this end, we use the mean and SD values of memorability scores in each category to quantify such relationship. As shown in Fig. \ref{fig:scenecat}, the horizontal axis represents scene categories in the descending order of the corresponding average memorability scores. The average score ranges from 0.79 to 0.36, giving a sense of how memorability changes across different scene categories. The distribution in Fig. \ref{fig:scenecat} indicates that some unusual classes like aurora tend to be more memorable, while usual classes like mountain are more likely to be forgotten. This is possibly due to the frequency of each category appears in daily life.


\section{Predicting outdoor natural scene memorability}

In above, we have analyzed how low-, middle- and high-level visual features affect the memorability of outdoor natural scenes. Now, we focus on the prediction of outdoor natural scene memorability in this section. Since DNN models have shown splendid performance in various computer vision tasks, in Section \ref{deep}, we first discover the effectiveness of the features extracted by DNN (i.e., deep features) on estimating the memorability of outdoor natural scenes. Then, in Section \ref{DeepNSM}, an end-to-end DNN method, called DeepNSM, is proposed to predict outdoor natural scene memorability.

\subsection{Deep features vs. memorability}\label{deep}


In recent years, DNN is utilized to predict generic image memorability \cite{khosla2015understanding,baveye2016deep,dubey2015makes}. For outdoor natural scene images, to dig out how deep features influence their memorability, we fine-tuned MemNet\footnote{MemNet is proposed to predict the memorability scores of generic images.} \cite{khosla2015understanding} on the training set of our LNSIM database, using the Euclidean distance between the predicted and ground truth memorability scores as the loss function. We extract the output of the last hidden layer as the deep features (dimension: 4096).
Due to the strong ability of DNN to extract spatial features, the learned deep features may consist of hierarchical features from low- to high-level.

To evaluate the correlation between the deep features and outdoor natural scene memorability, similar to above handcrafted features, an SVR predictor with histogram intersection kernel is trained for the deep features. The SRCC of deep features is 0.44 (p-value $=3.8547\times10^{-22}$), exceeding all handcrafted features. It is observed that DNN indeed works well on predicting the memorability of outdoor natural scenes, as the deep features show a rather high prediction accuracy. Nonetheless, there is no doubt that the fine-tuned MemNet also has its limitation, since it still has gap to human consistency ($\rho=0.78$).

We further combine the deep features with each of the aforementioned low-, middle- and high-level feature, to explore whether such combination is able to improve the prediction accuracy. The SRCC values of these combinations are shown in Table \ref{tab:deep}. Unfortunately, it can be seen that low- and middle-level features do not boost the SRCC of the deep features. It is probably because DNN has the ability to extract hierarchical features of different levels, leading to the ineffectiveness of low- and middle-level features. However, the scene category, as a high-level feature, helps to increase the SRCC of the deep features from $\rho=0.4406$ to $\rho=0.4610$.  This may be due to the fact that the architecture of MemNet is too simple to adequately learn the high-level feature, so that combining with the high-level feature is advantageous for the deep features to predict outdoor natural scene memorability. Motivated by this, we propose a scene category based DNN approach to predict the memorability of outdoor natural scene images in the next section.

\subsection{DeepNSM: DNN for outdoor natural scene memorability}\label{DeepNSM}

\begin{figure*}[t]
\begin{center}
\includegraphics[width=.97\linewidth]{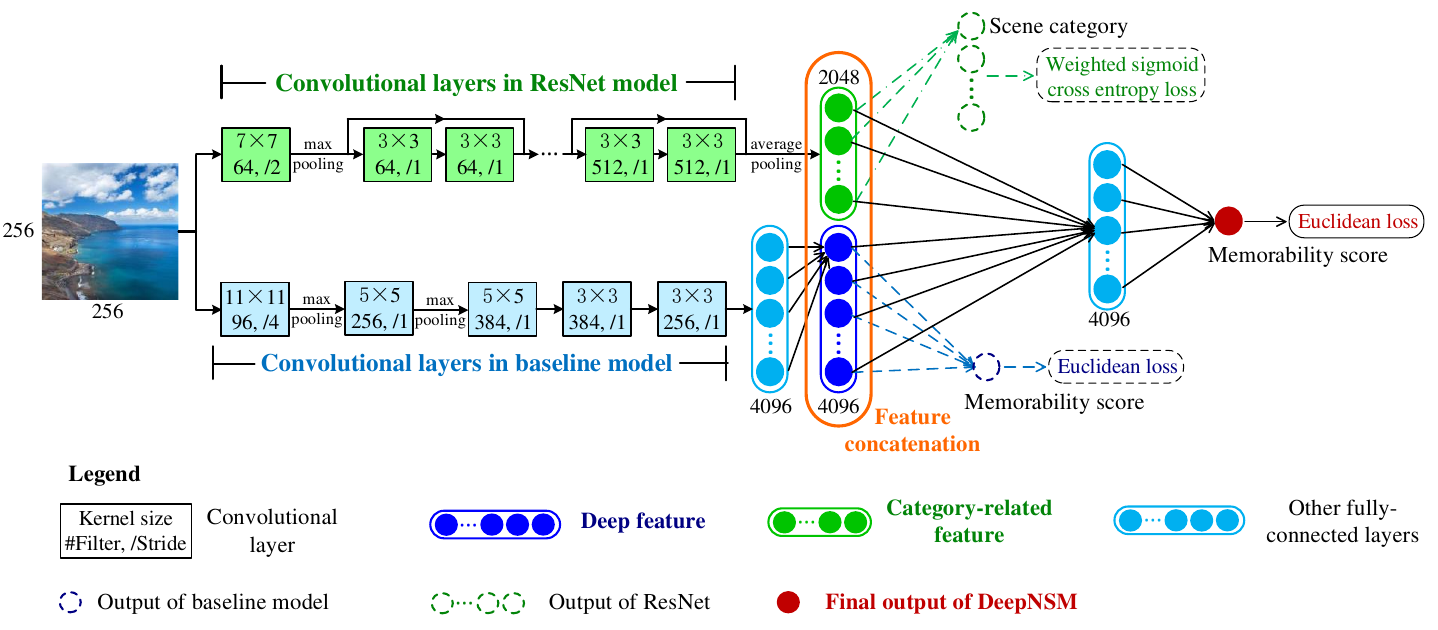}
\end{center}
\caption{Architecture of our DeepNSM model.}\label{fig:net}
\end{figure*}

As aforementioned, MemNet, which is fine-tuned on our training set, outperforms all the low-, middle- and high-level visual features. Hence, the fine-tuned MemNet model serves as the baseline model on predicting outdoor natural scene memorability.
Furthermore, according to the analysis above, the deep features combined with scene category are the most effective in predicting the memorability of outdoor natural scenes. Therefore, we propose an end-to-end DeepNSM method, which exploits both deep and category-related features for predicting the outdoor natural scene memorability.
Note that, in this paper, the ``deep features'' particularly refer to the 4096 output numbers of the last hidden layer in the baseline model.


\textbf{Extracting category-related features.} In DeepNSM, ResNet \cite{he2016deep} is applied to extract the category-related features. We first initialize ResNet with the pre-trained model on ImageNet \cite{Deng2009ImageNet}. Then, 33,000 outdoor natural scene images selected from the database of Places \cite{zhou2017places} are adopted to fine-tune the ResNet model. Finally, it is further fine-tuned on our training set according to the ground truth labels of scene category. Note that, different from the databases of ImageNet and Places, whose labels are one-hot, each image in our LNSIM database may contain the scenes of several categories. As such, it is a multi-label classification task. Thus, the weighted sigmoid cross entropy is utilized as the loss function, instead of softmax loss in \cite{he2016deep}. The fine-tuned ResNet can be seen as an extractor of category-related features. The output of the hidden fully-connected layer in ResNet is used as the extracted category-related features. See Fig. \ref{fig:net} for more details.

\textbf{The proposed architecture.} Finally, the architecture of our DeepNSM model is presented in Fig. \ref{fig:net}. In our DeepNSM model, the aforementioned category-related features are concatenated with the deep features obtained from the baseline model. Based on such concatenated features, additional fully-connected layers (including one hidden layer with dimension of 4096) are designed to predict the memorability scores of outdoor natural scene images.
Note that although some existing memorability prediction works \cite{isola2014makes,isola2011makes} also take image category into consideration, they only apply the manually classified ground truth category information. To the best of our knowledge, our work is the first attempt to automatically extract the category-related features by DNN in predicting memorability. The advantage is two fold: (1) The image memorability can be predicted without any manual annotation; (2) It is able to achieve the end-to-end training of the DNN model.

\section{Experiments}

\subsection{Settings}
The experimental results are presented to validate the effectiveness of our DeepNSM approach in predicting the memorability of outdoor natural scene images. Recall that all 2,632 outdoor natural scene images with their ground truth memorability scores in our LNSIM database introduced in Section \ref{data} are randomly divided into the training set (2,200 images) and the test set (432 images). When training the DeepNSM, the layers of the baseline and ResNet models are initialized by the individually pre-trained models, and the additional fully-connected layers are randomly initialized. The whole network is jointly trained in an end-to-end manner, using the Adam \cite{Kingma2014Adam} optimizer with the Euclidean distance adopted as the loss function.

\subsection{Performance evaluation}

Now, we evaluate the performance of our DeepNSM model on predicting outdoor natural scene memorability in terms of SRCC ($\rho$). Our DeepNSM model is tested on both the test set of our LNSIM database and the NSIM database introduced in \cite{Lu2016Predicting}. The SRCC performance of our DeepNSM model is compared with the state-of-the-art memorability prediction methods, including Isola \textit{et al.} \cite{isola2011makes,isola2014makes}, MemNet \cite{khosla2015understanding}, MemoNet \cite{baveye2016deep} and Lu \textit{et al.} \cite{Lu2016Predicting}\footnote{Some features of Isola \textit{et al.} \cite{isola2011makes,isola2014makes} are annotation labels which are hard to be obtained on other datasets, so we only use their reproducible features for the results of \cite{isola2011makes,isola2014makes, Lu2016Predicting} on our LNSIM database in Tables \ref{tab:results} and \ref{tab:mae} and for their results of the models re-trained by LNSIM in Tables \ref{tab:retrain} and \ref{tab:mae1}. Note that, the results of \cite{isola2011makes,isola2014makes, Lu2016Predicting} on the NSIM dataset~\cite{Lu2016Predicting} are fully re-implemented with all features in Tables \ref{tab:results} and \ref{tab:mae}, since NSIM is a part of the dataset in Isola \textit{et al.} \cite{isola2011makes,isola2014makes} with all features available.}. Among them, Isola \textit{et al.} \cite{isola2014makes}, MemNet \cite{khosla2015understanding} and MemoNet \cite{baveye2016deep} are the latest methods for generic images. Lu \textit{et al.} \cite{Lu2016Predicting} is a state-of-the-art method for predicting outdoor natural scene memorability.

\textbf{Comparison with latest generic methods.} Table \ref{tab:results} shows the SRCC performance of our DeepNSM and the three compared methods. Our DeepNSM successfully achieves the outstanding SRCC performance, i.e., $\rho=0.58$ and $0.55$, over the LNSIM and NSIM \cite{Lu2016Predicting} databases, respectively. It significantly outperforms \cite{isola2011makes,isola2014makes}, and also achieves better performance than the state-of-the-art DNN methods, MemNet \cite{khosla2015understanding} and MemoNet \cite{baveye2016deep}. Besides, the p-values of our DeepNSM method are $4.7045\times10^{-39}$ and $7.4548\times10^{-13}$ on the LNSIM and NSIM databases, respectively. This validates that our predicted memorability scores (p-value $\ll1.67\times 10^{-3}$) are statistically significant.

\begin{table}[!t]
\scriptsize
  \centering
    \caption{The SRCC ($\rho$) of our DeepNSM and compared methods.}\label{tab:results}
    \vspace{-1.2em}
    \begin{tabular}{cccccc}
    \toprule
       \multirow{2}[2]*{Database}&MemNet &MemoNet  &Lu \textit{et al.} & Isola \textit{et al.} &Our  \\
        & \cite{khosla2015understanding} & \cite{baveye2016deep} & \cite{Lu2016Predicting} &\cite{isola2011makes,isola2014makes} &DeepNSM \\
    \midrule
    Our LNSIM  &0.43  & 0.39  & 0.19 & 0.15 & \textbf{0.58} \\
    \midrule
    NSIM \cite{Lu2016Predicting}   & 0.40   & -*& 0.47  & 0.42  & \textbf{0.55} \\
    \bottomrule
   \\
    \multicolumn{6}{p{8.5cm}}{* MemoNet is not tested on the NSIM database, since the NSIM database is completely included in the training set of MemoNet.}\\
    \end{tabular}
\end{table}

\begin{table}[!t]
  \centering
  \scriptsize
    \caption{The SRCC ($\rho$) of our DeepNSM and compared methods re-trained by LNSIM training set.}\label{tab:retrain}
    \begin{tabular}{ccccccc}
    \toprule
      \multirow{2}[2]*{Database}&MemNet &MemoNet  &Lu \textit{et al.} & Isola \textit{et al.} &Our  \\
        & \cite{khosla2015understanding} & \cite{baveye2016deep} & \cite{Lu2016Predicting} &\cite{isola2011makes,isola2014makes} &DeepNSM \\
    \midrule
    Our LNSIM  &  0.50 & 0.42  &  0.33 & 0.33  &\textbf{0.58} \\
    \midrule
    NSIM \cite{Lu2016Predicting}   &  0.44  & 0.43 &  0.18  & 0.16 &   \textbf{0.55} \\
    \bottomrule
    \end{tabular}
\end{table}

The above results demonstrate the effectiveness of our DeepNSM model in predicting outdoor natural scene memorability. It is worth pointing out that as claimed in \cite{khosla2015understanding} and \cite{baveye2016deep}, both MemNet and MemoNet methods are able to reach $\rho=0.64$ on generic images. Nevertheless, their performance severely degrades on outdoor natural scenes, and thus validates the difference of factors influencing the memorability of generic and outdoor natural scene images. Besides, it also reflects the difficulty to accurately predict outdoor natural scene memorability. In summary, our DeepNSM model outperforms the state-of-the-art generic methods on predicting outdoor natural scene memorability, making up the shortcomings of these generic image methods.

\textbf{Comparison with the latest outdoor natural scene method.} We compare our DeepNSM model with the latest method \cite{Lu2016Predicting}, which is designed for predicting outdoor natural scene memorability. As shown in Table \ref{tab:results}, our DeepNSM model outperforms the method of \cite{Lu2016Predicting} on both LNSIM and NSIM databases. Moreover, compared with the database of NSIM \cite{Lu2016Predicting} ($\rho=0.47$), the SRCC of \cite{Lu2016Predicting} obviously reduces on our LNSIM database ($\rho=0.19$). On the contrary, our DeepNSM model achieves comparable performance on both databases. This shows the good generalization capacity of our DeepNSM model, which benefits from the large scale training set of our LNSIM database.

\textbf{Re-training the compared methods by LSNIM.}  For fair comparison, we also re-trained all compared methods over our LNSIM training set, making our DeepNSM and compared methods share the same training data. The SRCC ($\rho$) performance of our DeepNSM method and the re-trained compared methods are reported in Table \ref{tab:retrain}. It can be seen from Tables \ref{tab:results} and \ref{tab:retrain} that the performance of the four compared methods increase after re-training. However, our DeepNSM methods still significantly outperforms all other methods for predicting the memorability of outdoor natural scenes.

\begin{table}[!t]
  \centering
  \tiny
    \caption{The MAE / MSE of our DeepNSM and compared methods.}\label{tab:mae}
    \begin{tabular}{cccccc}
    \toprule
       \multirow{2}[2]*{Database}&MemNet &MemoNet  &Lu \textit{et al.} & Isola \textit{et al.} &Our  \\
        & \cite{khosla2015understanding} & \cite{baveye2016deep} & \cite{Lu2016Predicting} &\cite{isola2011makes,isola2014makes} &DeepNSM \\
    \midrule
    Our LNSIM  &  0.1081\ /\ 0.0183 &  0.1152\ /\ 0.0211 & 0.1206\ /\ 0.0220 & 0.1369\ /\ 0.0294 & \textbf{0.0998\ /\ 0.0153}\\
    \midrule
    NSIM \cite{Lu2016Predicting}   & 0.1093\ /\ 0.0199   & -* &  0.1110\ /\ 0.0187  & 0.1121\ /\ 0.0191  & \textbf{0.0916\ /\ 0.0145} \\
    \bottomrule
   \\
    \multicolumn{6}{p{8.5cm}}{* MemoNet is not tested on the NSIM database, since the NSIM database is completely included in the training set of MemoNet.}\\
    \end{tabular}
\end{table}

\begin{table}[!t]
  \centering
  \tiny
    \caption{The MAE / MSE of our DeepNSM and compared methods re-trained by LNSIM database.}\label{tab:mae1}
    \begin{tabular}{cccccc}
    \toprule
       \multirow{2}[2]*{Database}&MemNet &MemoNet  &Lu \textit{et al.} & Isola \textit{et al.} &Our  \\
        & \cite{khosla2015understanding} & \cite{baveye2016deep} & \cite{Lu2016Predicting} &\cite{isola2011makes,isola2014makes} &DeepNSM \\
    \midrule
    Our LNSIM  &  0.1098\ /\ 0.0191  &   0.1090\ /\ 0.0186  &   0.1133\ /\ 0.0199 & 0.1140\ /\ 0.0201 & \textbf{0.0998\ /\ 0.0153}\\
    \midrule
    NSIM \cite{Lu2016Predicting}   & 0.1045\ /\ 0.0192& 0.1028\ /\ 0.0176 & 0.1696\ /\ 0.0422 & 0.1676 \ /\ 0.0414 & \textbf{0.0916\ /\ 0.0145} \\
    \bottomrule

    \end{tabular}
\end{table}
\textbf{Performance in terms of mean absolute error (MAE) and mean square error (MSE).} Moreover, we also evaluate the prediction accuracy of our DeepNSM approach and the compared approaches in terms of the MAE and MSE between the predicted memorability scores and the ground-truth. The results are shown in Tables \ref{tab:mae} and \ref{tab:mae1}. Note that, in Table \ref{tab:mae}, the compared methods are trained on their original training data. In Table \ref{tab:mae1}, we re-train all compared methods on our LNSIM database. As Tables \ref{tab:mae} and \ref{tab:mae1} show, our DeepNSM approach achieves MAE = 0.0998 and MSE = 0.0153 on our LNISM test set, both lower than all compared methods. Besides, on the NSIM test set \cite{Lu2016Predicting}, our approach also outperforms all compared methods in terms of MAE and MSE. These validates that our DeepNSM approach has the highest accuracy for predicting both the memorability score (MAE/MSE) and memorability rank ($\rho$) of outdoor natural scenes.

\subsection{Ablation analysis}

In ablation experiments, we analyze the performance of our category-feature extractor, our baseline model and the improvement of combining category-related features.

\textbf{Ablation studies on scene classification.} On our LNSIM test set, the fine-tuned ResNet has the mean average precision (mAP) of 70.63\% for the multi-label scene classification, thus verifying the effectiveness of our category-feature extractor. Besides, we also evaluate the performance of the solely category feature, by replacing the ground-truth label with the estimated category. It has $\rho=0.35$, which is only slightly lower than applying the ground-truth labels ($\rho=0.38$ in Section IV-C). This can be seen as the baseline performance of the category feature, without using human annotations.

\textbf{Ablation studies of our DeepNSM model.} Then, the SRCC of our baseline model on the test set of LNSIM database reaches $\rho=0.50$, higher than all three compared methods. Hence, our baseline model serves as a solid cornerstone to predict outdoor natural scene memorability. As Table \ref{tab:results} shows, after combining the category-related feature, the performance of our DeepNSM model increases to $\rho=0.58$. It convincingly verifies the effectiveness of scene category on outdoor natural scene memorability prediction. Furthermore, as discussed in Section \ref{deep}, the SVR predictor trained by the 4096-dimension deep feature of the baseline model yields $\rho=0.44$. Adding scene category feature to the SVR predictor only slightly improves the SRCC to $\rho=0.46$ ($\Delta\rho=0.46-0.44=0.02$). However, taking advantage of DNN, the SRCC increase is significantly enlarged ($\Delta\rho=0.58-0.50=0.08$) in our DeepNSM model, when concatenating category-related feature extracted from ResNet.
This shows the remarkable ability of our DeepNSM model in learning to predict outdoor natural scene memorability from the above concatenated features.

\begin{figure*}[t]
\vspace{-1em}
\begin{center}
\includegraphics[width=.96\linewidth]{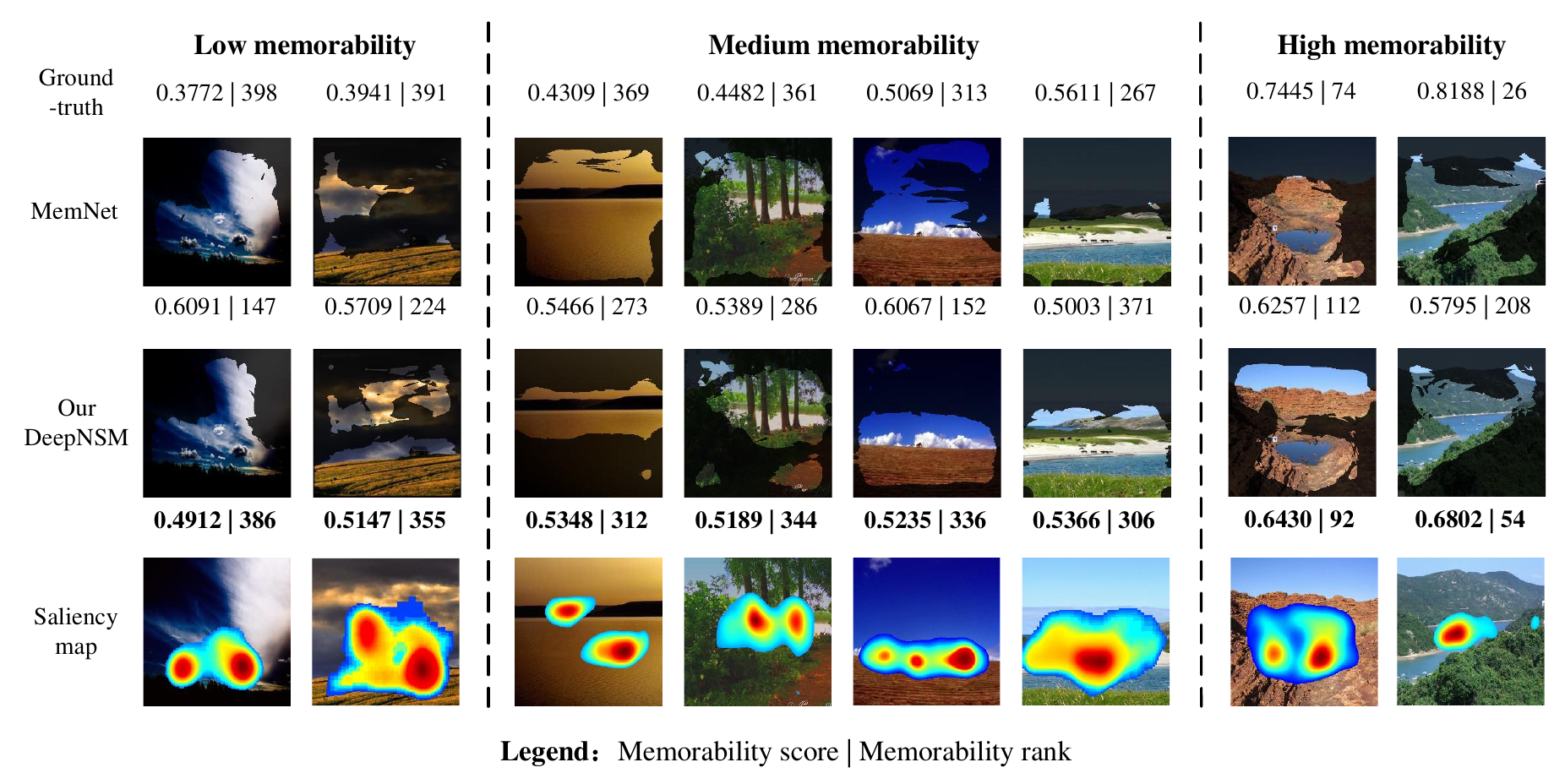}
\end{center}
\vspace{-1em}
\caption{Feature visualization of the last convolutional layer and the predicted saliency maps.}\label{fig:vis}
\end{figure*}

\textbf{Parameter number and time complexity.} Our baseline model has 62.4 M parameters, and consumes 0.0056 second to predict the memorability score of one image on the computer with one GeForce GTX TITAN X GPU. After combining the category-feature extractor, the parameter number of our DeepNSM model increases to 105.5 M, which is 1.69 times of the baseline model. Besides, our DeepNSM model takes 0.0296 second to calculate the memorability score of one image. That is, our DeepNSM approach is able to predict the memorability of more than 33 images in one second.


\subsection{Understanding the memorability and our DeepNSM model}

In this section, we first focus on understanding the memorability of outdoor natural scenes. That is, we analyze the reason why scene category is an effective feature for memorability prediction and what makes an outdoor natural scene image memorable. Then, we investigate and visualize the internal representation learned by our DeepNSM model to understand how our DeepNSM model works to predict memorability.

\begin{table}[t]
  \centering
  \footnotesize
    \caption{The word frequencies ($\%$) of scene categories.}\label{tab:fre}
    \begin{tabular}{|c|c|c|c|}
    \hline
    & High memo.&Medium memo.&Low memo.\\
    \hline
     Memorability rank & 1-20 & 21-50 & 51-71 \\
    \hline
    Average frequency  &  $4.036\!\!\times\!\!10^{-4}$ & $13.589\!\!\times\!\!10^{-4}$ & $10.313\!\!\times\!\!10^{-4}$ \\
    \hline
    \end{tabular}
    \vspace{-1em}
\end{table}

\textbf{Scene category.} Brown \textit{et al.}, \cite{brown1977memorability} investigated that the memorability of items has relationship with word frequency. The memorability of outdoor natural scenes may be also related to the frequency of scenes to appear in daily life. Therefore, we make statistics of the word frequencies for the 71 scene categories shown in Fig. \ref{fig:scenecat} using Google Books Ngram Viewer\footnote{\url{https://books.google.com/ngrams/}}. We follow the Fig. \ref{fig:scenecat} to rank the categories in descending order of their memorability, and the average word frequencies are shown in Table \ref{tab:fre}. We can see from Table \ref{tab:fre} that the 20 categories of the highest memorability scores occur with the least frequency of $4.036\times10^{-4}\%$. That is, the scenes that rarely occurs are easiest to be memorized.
For example, in Fig. \ref{fig:Figure 1}, aurora is uncommon (word frequency = $0.337\times10^{-4}\%$) compared with desert (word frequency = $17.672\times10^{-4}\%$), lake (word frequency = $20.747\times10^{-4}\%$), etc., and it is also rarely seen in daily life, making it more memorable than other three categories.
In conclusion, the effectiveness of the scene category feature for predicting memorability is probably because that different categories of scenes appear in human life with different frequency. As a result, the outdoor natural scenes which appear with low frequency do not collide with previous similar memory, and thus they are easy to be memorized by human brain.
Hence, when designing the magazine covers or posters, it is helpful to use images with the natural scenes that are infrequently seen in daily life, for leaving deep impression in people's mind.


\begin{figure*}[!t]
\begin{center}
\includegraphics[width=.95\linewidth]{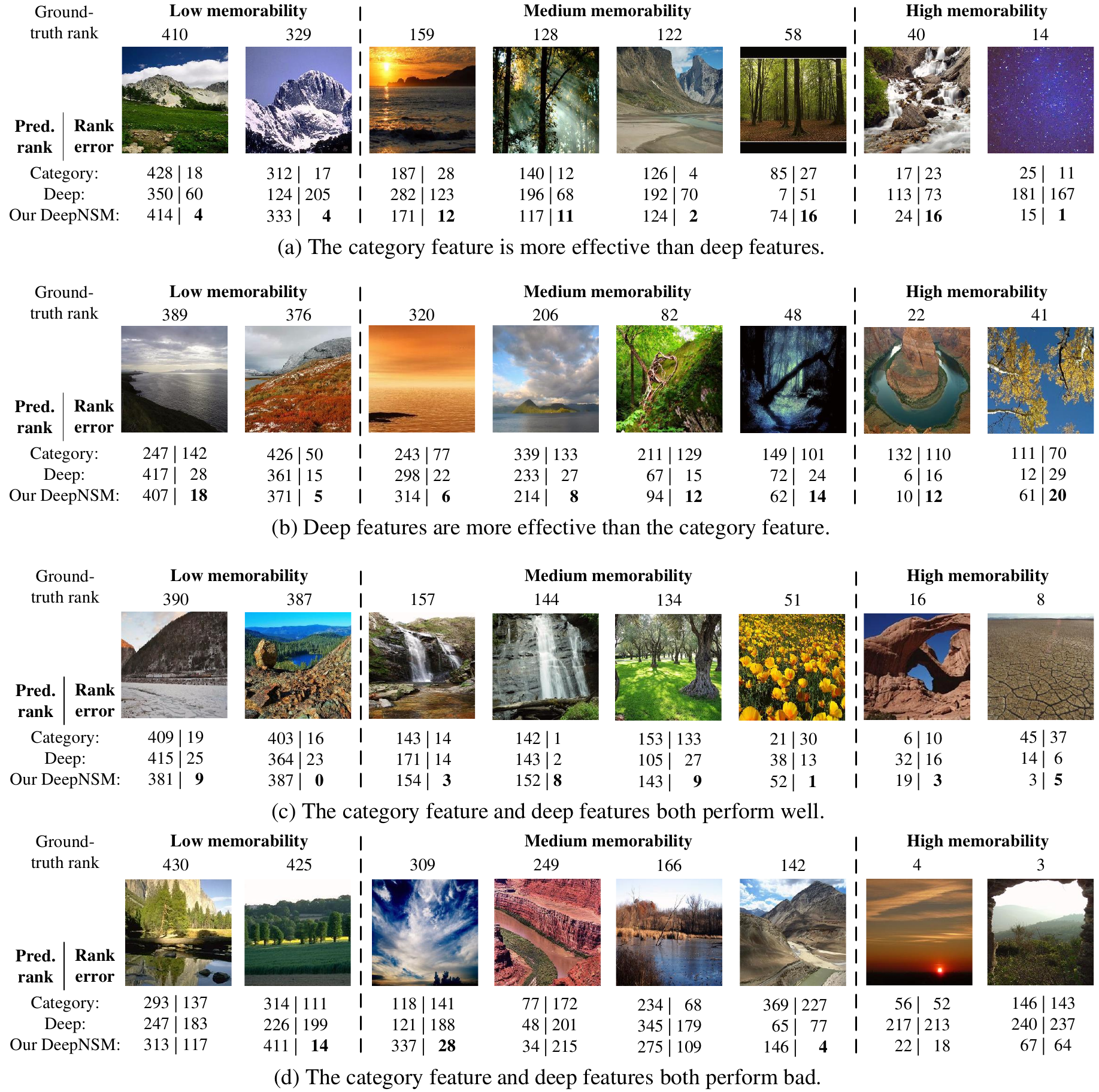}
\end{center}
\caption{Example performances of the category feature, deep features and our DeepNSM model. }\label{fig:case}
\end{figure*}

\textbf{Visualization of DNN models.} We investigate the internal representation learned by DeepNSM, and we set MemNet as the baseline. We apply a data-driven receptive field method \cite{bolei2015object} to visualize segmentation produced by the last convolutional layer of both MemNet and DeepNSM, and utilize the saliency detection method SALICON \cite{jiang2015salicon} to predict saliency maps. Fig. \ref{fig:vis} shows the visualization results, where the light regions make main contribution to memorability prediction of outdoor natural scenes. We observe that light regions are highly correlated with whether the image is easy to be memorized or not. Compared with Memnet, our DeepNSM approach is learned to allocate memory regions better. For example, in the second image of the low memorability group, in accordance with the saliency map, the orange clouds and the sky in the middle of the images are likely to attract people attention when looking through this image. The MemNet, designed for generic images, fails to predict memorability depending on this region. In contrast, our DeepNSM model successfully learns to focus on the orange clouds, leading to better perdition result.
In the third image of medium memorability group, according to the saliency prediction results, people are likely to pay more attention on the white cloud and the surrounded ground and sky. It can be seen from Fig. \ref{fig:vis}, our DeepNSM model locates the memory region more precisely then MemNet, and therefore results in better performance for predicting memorability. In conclusion, our DeepNSM model is able to locate memorability region more correctly and precisely than the generic method MemNet, and achieves better performance on images from low-, medium-, to high-memorability.

\subsection{Case study}

Finally, as illustrated in Fig. \ref{fig:case}, we study the successful and failure cases of the category feature, the deep features of baseline DNN model and our DeepNSM model. As Fig. \ref{fig:case} shows, we analyze the following four cases: (a) the category feature is effective while deep features are ineffective; (b) deep features are effective while the category feature is ineffective; (c) they both perform well; (d) they are both not effective. Note that, in Figure \ref{fig:case}, we use the absolute error between the ground-truth and the predicted memorability ranks to evaluate the effectiveness of each feature/method. If the rank error is less than $10\%$ of the total number of images, i.e., $10\%\times432\approx43$, the feature/method can be seen as effective. The number above each image is the ground-truth memorability rank, and the left column numbers under images are the predicted ranks, and then the right column numbers under images are the rank errors.

Specifically, Figure \ref{fig:case}-(a) shows the examples that the category feature, as a non-deep feature, is effective while deep features are not. Fortunately, our DeepNSM model takes advantage of both deep features and the category-related feature, thus still achieving good performance on memorability prediction. Similarly, in Figure \ref{fig:case}-(b), deep features are effective while the category feature does not work. Our DeepNSM model is also able to effectively predict the memorability, thanks to using both deep and category features. Figure \ref{fig:case}-(c) demonstrates examples that both deep and category features performs well, leading to the good performance of our DeepNSM model. Finally, Figure \ref{fig:case}-(d) shows the images that the category and deep features both fail to accurately predict the memorability. We can see from Figure \ref{fig:case}-(d) that our DeepNSM model still performs well on some images. This is probably because the solely category feature or deep features cannot represent the memorability well on these images, but the effective features for memorability prediction is able to be learnt from their combination. However, there are also some examples shown in Figure-(d), in which our DeepNSM model is not effective, because of the bad performances of category and deep features. This shows that although our DeepNSM model outperforms all previous methods, there still exists some cases that we are not able to accurately predict the memorability.




We first investigate the prediction error of our DeepNSM method on images with low- ($\rho<0.4$), medium- ($0.4\leq\rho<0.7$) and high-memorability ($\rho\geq0.7$). Here, we use the absolute ranking error between predicted and ground-truth memorability scores to evaluate the prediction error. The experimental results show that the rank error is 81.9 averaged on all low-memorability images, and this figure is 90.8 and 84.1 for medium- and high-memorability images, respectively. This indicates that our DeepNSM methods predict memorability with higher accuracy on low- and high-memorability images, and with worse performance for medium-memorability images.

\begin{table}[t]
  \centering
  \footnotesize
  \caption{Analysis of the GLCM features and the ranking errors}
    \begin{tabular}{|c|l|r|r|r|r|}
    \hline
    \multicolumn{2}{|c|}{Image groups} & \multicolumn{1}{c|}{Group 1} & \multicolumn{1}{c|}{Group 2} & \multicolumn{1}{c|}{Group 3} & \multicolumn{1}{c|}{Group 4} \\
    \hline
    \hline
    \multicolumn{1}{|c|}{\multirow{3}[1]{*}{\tabincell{c}{GLCM \\ features}}} & Contrast & 1196.9 & 487.2 & 247.6 & 89.2\\
\cline{2-6}          & Homogeneity & 0.1338 & 0.2157 & 0.2875 & 0.4081 \\
\cline{2-6}          & Correlation & 0.8022 & 0.9119 & 0.9487 & 0.9676 \\
    \hline
    \hline
    Average & Category & 123.8 & 103.6 & \textbf{102.0} & 114.3 \\
\cline{2-6}   ranking       & Baseline DNN & \textbf{87.4} & 107.9 & 114.2 & 106.1 \\
\cline{2-6}     errors     & our DeepNSM & \textbf{79.5} & 85.4  & 89.0    & 99.7 \\
    \hline
    \end{tabular}%
  \label{tab:case}%
\end{table}%

Then, we also utilize the gray-level co-occurrence matrix (GLCM) for case studies. Specifically, we use the contrast, homogeneity and correlation of GLCM to investigate the relationship between the image texture and the prediction errors of the non-deep feature (i.e., scene category), deep features (i.e., the baseline DNN) and our DeepNSM model. As shown in Table \ref{tab:case}, we sort the 432 test images from high to low contrast values and split them into four groups, i.e., the images of Group 1 are with the highest contrast, and Group 4 contains the lowest contrast images. Interestingly, the average homogeneity and correlation values increase while the contrast reduces from Group 1 to Group 4. This indicates the images of Group 1 are with the highest texture complexity and the complexity decreases alongside Group 1 to Group 4.

Table \ref{tab:case} tabulates the average ranking errors of the category feature, deep features and our DeepNSM model in each group. It can be seen from Table XI that the category feature is with the lowest ranking error on Groups 2 and 3, while the error of deep features is higher on them. However, the deep features have obviously lower prediction error than the category feature for Group 1. This indicates that the non-deep feature of scene category performs well on images with medium texture complexity, and the deep features achieve the best performance on images with high texture complexity. Therefore, the complementary of the category and deep features leads to the better performance when combining them together in our DeepNSM model.

More importantly, our DeepNSM model outperforms both category and deep features on all groups. This shows that integrating the category-related features to the baseline DNN in our DeepNSM model improves the prediction accuracy of outdoor natural scene memorability, validating the effectiveness of our DeepNSM model. Besides, it can be seen from the final performance of our DeepNSM model that the memorability of images with higher contrast, lower homogeneity and lower intra-correlation are more likely to be accurately predicted by our DeepNSM model. This is probably because these  images  are  with  more information  and  more  obvious spatial features, which facilitate our DeepNSM model to extract effective features for predicting the memorability.

\subsection{Future works}

First, although our LNSIM database is currently the largest one for the memorability of outdoor natural scenes, it can be further enlarged in the future to include more various kinds of scene categories and ensure the sufficient number of images in each category. Besides, as shown in Table XI, for images with low contrast, high homogeneity and intra-correlation, our DeepNSM model fails to capture effective features. Therefore, incorporating more features, which affect the memorability of outdoor natural scenes, is probably a promising research direction to break through the failure cases of our DeepNSM model.

Furthermore, although the prediction accuracy of generic images can be boosted by utilizing our DeepNSM model on the sub-set images of outdoor natural scene, the current DeepNSM model cannot be directly applied for improving the performance on predicting the memorability of generic images. However, the study on outdoor natural scenes may help to understand the contribution of natural scene background (as shown in Fig. \ref{fig:new1}) to the memorability of a whole image. Hence, it is indeed an interesting future work that takes advantage of the study on the memorability of outdoor natural scenes to improve the performance on predicting generic image memorability.

\section{Conclusion}
In this paper, we have investigated the memorability of outdoor natural scene from data-driven perspective. Specifically, we established the LNSIM database that helps to study and analyze the human memorability on outdoor natural scene in depth. In exploring the correlation of memorability with low-, middle- and high-level features, we found that high-level feature of scene category plays an important role in predicting the memorability of outdoor natural scene. In addition, deep features show a positive impact on promoting the prediction performance on outdoor natural scenes. Accordingly, we proposed the DeepNSM method for predicting outdoor natural scene memorability in an end-to-end manner. The experimental results showed that our DeepNSM model advances the state-of-the-art in memorability prediction of outdoor natural scene images. Then, we also tried to understand why the scene category is correlated to memorability, and how our DeepNSM model works to effectively predict the memorability of outdoor natural scenes. These help to deeply understand the memorability and our DeepNSM model. Moreover, we studied the effective and ineffective cases of the DeepNSM model, and concluded the possible future works.


\ifCLASSOPTIONcaptionsoff
  \newpage
\fi

\bibliographystyle{IEEEtran}
\bibliography{IEEEabrv}

\begin{IEEEbiography}[{\includegraphics[width=1\linewidth]{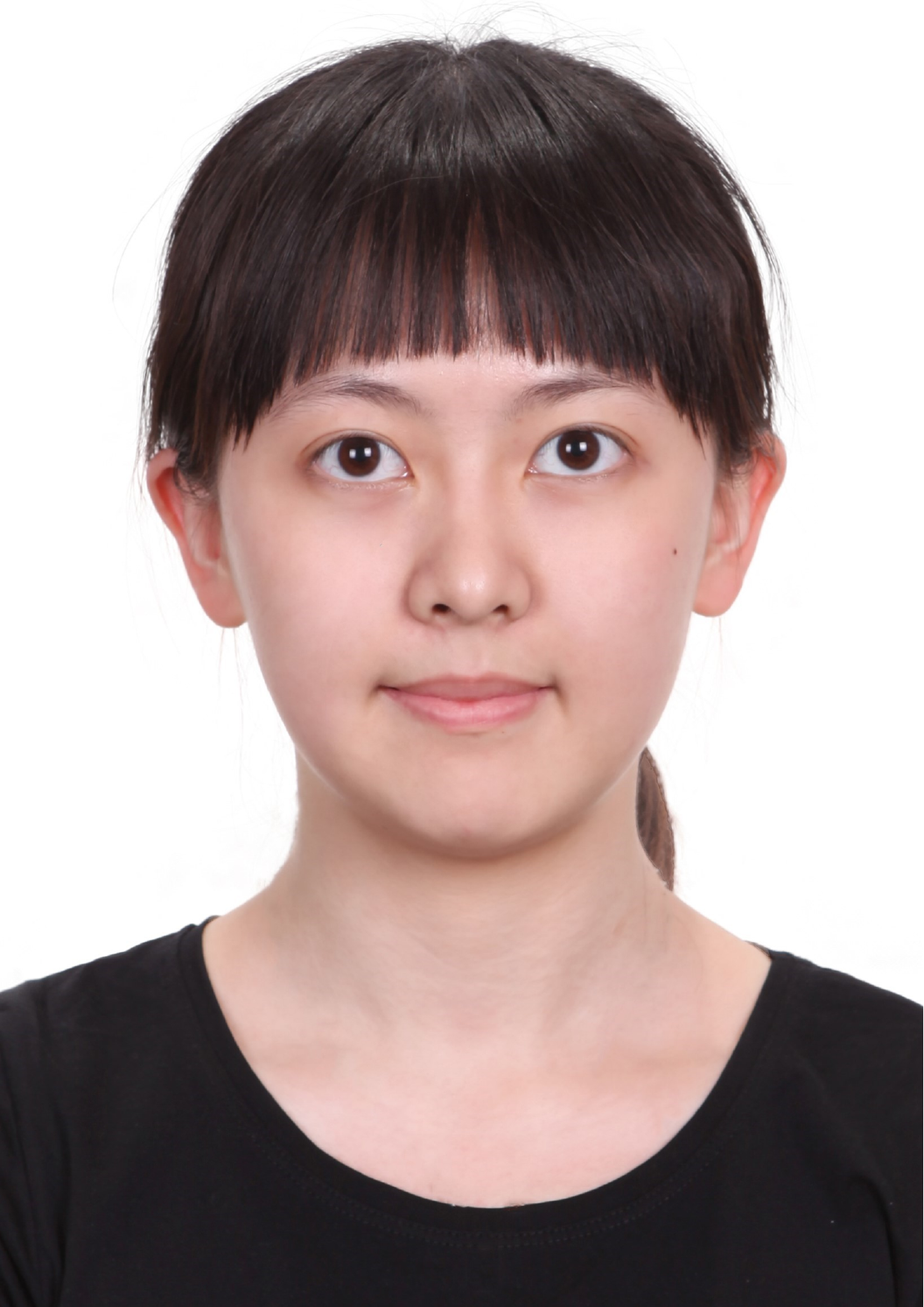}}]{Jiaxin Lu}
obtained the B.S. degree and the M.S. degree at the School of Electronic and Information Engineering, Beihang University in 2016 and 2019, respectively. Her research interests include computer vision and image processing. She has published several papers in international conference proceedings, e.g., IEEE International Conference on Visual Communications and Image Processing (VCIP).
\end{IEEEbiography}

\begin{IEEEbiography}[{\includegraphics[width=1\linewidth]{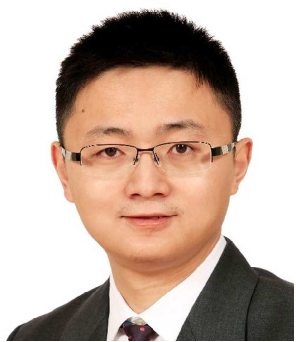}}]{Mai Xu}
(M'10, SM'16) received B.S. degree from Beihang University in 2003, M.S. degree from Tsinghua University in 2006 and Ph.D degree from Imperial College London in 2010. From 2010-2012, he was working as a research fellow at Electrical Engineering Department, Tsinghua University. Since Jan. 2013, he has been with Beihang University as an Associate Professor. During 2014 to 2015, he was a visiting researcher of MSRA. His research interests mainly include image processing and computer vision.  He has published more than 60 technical papers in international journals and conference proceedings, e.g., IEEE TIP, CVPR and ICCV. He is the recipient of best paper awards of two IEEE conferences.
\end{IEEEbiography}

\begin{IEEEbiography}[{\includegraphics[width=1\linewidth]{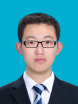}}]{Ren Yang}
received the B.S. degree from Beihang University in 2016, and obtained the M.S. degree at the School of Electronic and Information Engineering, Beihang University in 2019. His research interests mainly include computer vision and video coding. He has published several papers in international journals and conference proceedings, e.g., IEEE Transactions on Circuits and Systems for Video Technology, IEEE International Conference on Computer Vision and Pattern Recognition (CVPR) and IEEE International Conference on Multimedia \& Expo (ICME).
\end{IEEEbiography}

\begin{IEEEbiography}[{\includegraphics[width=1\linewidth]{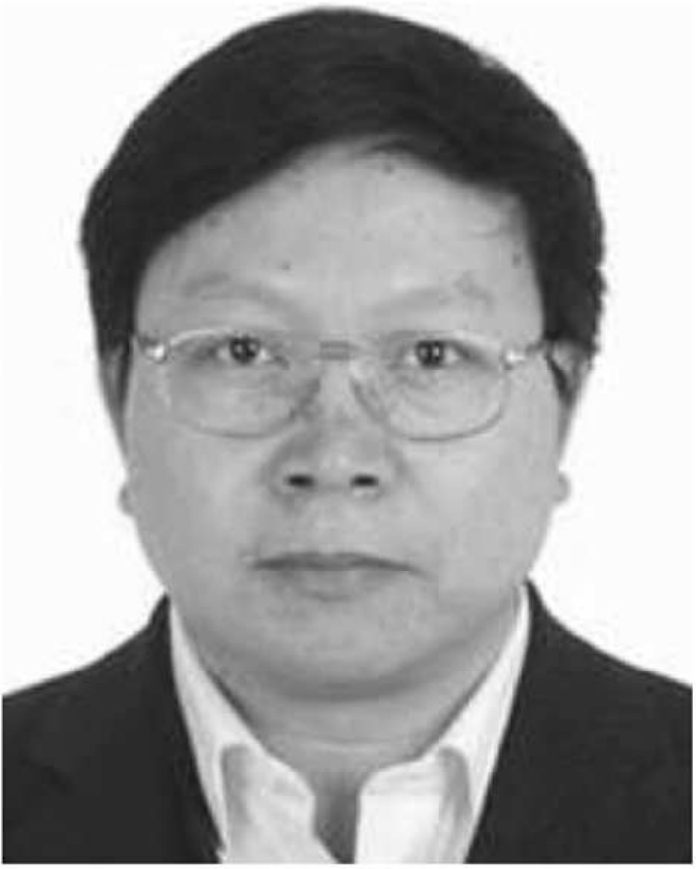}}]{Zulin Wang}
(M'14) received the B.S. and M.S. degrees in electronic engineering from Beihang University, in 1986 and 1989, respectively. He received his Ph.D. degree at the same university in 2000. His research interests include image processing, electromagnetic countermeasure, and satellite communication technology. He is author or co-author of over 100 papers and holds 6 patents, as well as published 2 books in these fields. He has undertaken approximately 30 projects related to image/video coding, image processing, etc.
\end{IEEEbiography}

\end{document}